\documentclass{article}
\usepackage{spconf,epsfig}
\usepackage{amsmath,amsfonts,amssymb,mathrsfs}
 \usepackage{booktabs}
 \usepackage{makecell}
   \graphicspath{   {images/}
                    {images/cuprite_result}
                    {images/MTRDR_result}
                    }
\usepackage[caption=false]{subfig}
   \DeclareGraphicsExtensions{.pdf,.jpeg,.png,.jpg}

\title{Weighted Sum of Segmented Correlation: An Efficient Method for Spectra Matching in Hyperspectral Images}
\name{
    Sampriti~Soor\textsuperscript{1}, Priyanka~Kumari\textsuperscript{2}, B.~S.~Daya~Sagar\textsuperscript{3}, Amba~Shetty\textsuperscript{2}
    \thanks{
        Codes: \texttt{\scriptsize https://github.com/SampritiSoor/WSSC} \newline
        e-mail ids: SS - \texttt{sampreetiworkid@gmail.com}, \newline PK - \texttt{singh.priyanka854@gmail.com},\newline BSDS - \texttt{bsdsagar@isibang.ac.in}, AS - \texttt{amba@nitk.edu.in}
    \newline
        Accepted in IEEE IGARSS 2024 conference
    }
}
\address{
    \textsuperscript{1}School of Computer Engineering, Kalinga Institute of Industrial Technology, Bhubaneswar, India
    \\
    \textsuperscript{2}Department of Water Resources and Ocean Engineering, NIT Karnataka
    Surathkal, India
    \\
    \textsuperscript{3}Systems Science and Informatics Unit, Indian Statistical Institute, Bangalore, India
}
\begin{document}
%
\maketitle
%
\begin{abstract}
Matching a target spectrum with known spectra in a spectral library is a common method for material identification in hyperspectral imaging research. Hyperspectral spectra exhibit precise absorption features across different wavelength segments, and the unique shapes and positions of these absorptions create distinct spectral signatures for each material, aiding in their identification. Therefore, only the specific positions can be considered for material identification. This study introduces the Weighted Sum of Segmented Correlation method, which calculates correlation indices between various segments of a library and a test spectrum, and derives a matching index, favoring positive correlations and penalizing negative correlations using assigned weights. The effectiveness of this approach is evaluated for mineral identification in hyperspectral images from both Earth and Martian surfaces.
\end{abstract}
%
\section{Introduction}
\label{sec:intro}
Hyperspectral imaging systems are capable of discerning subtle variations in reflectance spectra facilitating distinct spectral signatures. These signatures are unique patterns of reflectance values across diverse wavelengths influenced by the material's compositional attributes, in the form of absorptions that appear at distinct positions on the spectral domain. Absorptions are characterized by their distinctive shapes and depths. It is the combination of these absorption features within a spectrum that renders materials distinguishable, forming the foundation for precise material characterization and identification in hyperspectral imaging research.

Spectra are commonly identified by comparing their absorption signatures with known materials in a spectral library, traditionally done manually but increasingly facilitated by measures like cosine similarity, Gaussian likelihood, and correlation coefficient. Intrinsic shape-matching approaches, as outlined in \cite{clark2003imaging}, further enhance the performance of automated material identification processes. Since material identification primarily relies on the positions and shapes of absorption features in spectra, a simple distance-based similarity matching considering the entire spectral domain is inadequate. Instead, an approach that focuses on the most informative parts of a spectrum, particularly matching its distinguishable components, can significantly improve accuracy.
Real hyperspectral data often contains a spectrum that combines spectral information from multiple surface materials. This combined spectrum reflects the signatures of all the individual materials present. Although the individual signatures in a mixed spectrum may be suppressed or slightly altered in shape, they can generally be identified by examining their corresponding positions. Therefore, even in the case of mixed spectra, instead of matching the spectra using the entire spectral domain, it is more effective to identify contributing materials by comparing different segments of the spectrum.

In this paper, a spectra-matching technique is introduced that calculates correlation indices between different segments of a test spectrum and a library spectrum, and calculates a unified matching index. The process is described in detail in section 2 drawing a comparison with the related spectra-matching method described in \cite{clark2003imaging}. Section 3 includes the experimental results of the proposed method for identifying minerals in AVIRIS and CRISM MTRDR data. The results are visually compared with the results obtained by cosine similarity and correlation coefficient on the full spectral domain in section 3, whereas in section 4 a quantitative comparison is provided to establish the efficacy of the segmented matching approach. Finally, in Section 5, the key findings of this study are summarized.
\section{Weighted Sum of Segmented Correlation}
\label{sec:WSSC}
The section initially explores the computation of the correlation index between segments of a test spectrum and a library spectrum using the shape-matching algorithm outlined in \cite{clark2003imaging}. Subsequently, it addresses a limitation in this approach, establishes a link between the correlation index and Pearson's correlation coefficient, and streamlines the concept, presenting the proposed method for determining segmented correlation. The next step involves combining the segmented correlations of various sections to formulate the matching index through the Weighted Sum of Segmented Correlation.\par
\vspace{.1cm}
{\bf Correlation Index in \cite{clark2003imaging}:}\par
Let $L$ and $T$ respectively be the library and spectra segments on a wavelength segment $W$. Now, one can be most matched with the other by a translation involving shifting towards the other and enhancing/reducing its curvature. For $T$, the most effective translation to match with $L$ is given by eq. \ref{eq:LiTi}, while for $L$ to match with $T$ the same is given by eq. \ref{eq:TiLi}.\par
\begin{equation}\label{eq:LiTi}
    L=a_1+b_1T
\end{equation}
\begin{equation}\label{eq:TiLi}
    T=a_2+b_2L 
\end{equation}
where the coefficients $a_1$ and $a_2$ adjust the shift of the spectra, and $b_1$ and $b_2$ to modify the curvature during the translation process. The values of these coefficients are derived minimizing the quadratic errors.\par\noindent
From eq. \ref{eq:LiTi} 
\begin{equation}
\begin{aligned}
    \frac{\partial}{\partial a_1}(\sum(L - a_1 - b_1T )^2)&=0\\
    \frac{\partial}{\partial b_1}(\sum(L - a_1 - b_1T )^2)&=0
\end{aligned}
\end{equation}
solving which $b_1$ and $a_1$ is derived as
\begin{equation}\label{eq:a1b1}
\begin{aligned}
    b_1&=\frac{\sum(LT)-n\overline{L}\,\overline{T}}{\sum T^2-n\overline{T}^2}\\
    a_1&=\overline{L}-b_2\overline{T}\\
\end{aligned}
\end{equation}
Similarly from eq. \ref{eq:TiLi} $b_2$ and $a_2$ can be derived as
\begin{equation}\label{eq:a2b2}
    \begin{aligned}
    b_2&=\frac{\sum(LT)-n\overline{L}\,\overline{T}}{\sum L^2-n\overline{L}^2}\\
    a_2&=\overline{T}-b_1\overline{L}\\
    \end{aligned}
\end{equation}
The correlation index, $c$, is the geometric mean of $b_1$ and $b_2$, i.e.,
\begin{equation}
c=\sqrt{b_1b_2}=\frac{\sqrt{(\sum(LT)-n\overline{L}\,\overline{T})^2}}{\sqrt{(\sum T^2-n\overline{T}^2)(\sum L^2-n\overline{L}^2)}}
\end{equation}
where $n$ is the size of the segment.\par
\vspace{.1cm}
This correlation index is valued the same as the correlation coefficient of $L$ and $T$ but without the sign to indicate a positive or negative correlation\footnotemark.
In other words, this index determines the linear relationship between $L$ and $T$ correctly only if both change together.\par
\vspace{.1cm}
{\bf Proposed Correlation Index:}\par
\footnotetext{
The correlation coefficient between $L$ and $T$ is $\frac{Cov(L,T)}{\sigma_L\sigma_T}$\par
$=\frac{\frac{1}{n}\sum (L-\overline{L})(T-\overline{T})}{\sqrt{\frac{\sum(L-\overline{L})^2}{n}}\sqrt{\frac{\sum(T-\overline{T})^2}{n}}}$\par
$=\frac{\sum (LT)-\overline{L}\sum{T}-\overline{T}\sum{L}+\overline{L}\overline{T}\sum 1}{\sqrt{(\sum L^2-2\overline{L}\sum{L}+\overline{L}^2\sum 1)(\sum T^2-2\overline{T}\sum{T}+\overline{T}^2\sum 1)}}$\par
$=\frac{\sum(LT)-2n\overline{L}\overline{T}+n\overline{L}\overline{T}}{\sqrt{(\sum L^2-2n\overline{L}^2+n\overline{L}^2)(\sum T^2-2n\overline{T}^2+n\overline{T}^2)}}$\par
$=\frac{(\sum(LT)-n\overline{L}\,\overline{T})}{\sqrt{(\sum L^2-n\overline{L}^2)(\sum T^2-n\overline{T}^2)}}$
}
In the proposed approach the correlation index is calculated following the general expression of the correlation coefficient\footnotemark[\value{footnote}], so that it can identify both the positive and negative linear relation between $L$ and $T$. A positive correlation index positively influences the matching index, while a negative correlation index penalizes it. To improve the interpretability and simplify the calculation, both the spectra segments $L$ and $T$ are normalized to $L^\prime$ and $T^\prime$ to set the respective mean at 0 and standard deviation at 1. Thus the correlation coefficient can be derived as their co-variance and is calculated as
\begin{equation}
    c=\frac{1}{n}\sum (L^\prime T^\prime)
\end{equation}
Note that, if $L^\prime$ and $T^\prime$ are used in eq. \ref{eq:a1b1} and \ref{eq:a2b2}, both $a_1$ and $a_2$ are derived as 0, and both $b_1$ and $b_2$ are derived same as the proposed correlation index $\frac{1}{n}\sum (L^\prime T^\prime)$, because $\overline{L^\prime}=\overline{T^\prime}=0$ and $\sum L^{\prime\,2}=\sum T^{\prime\,2}=n$.\par
\vspace{.1cm}
{\bf Calculation of Matching Index:}\par
The computation of the matching index can be flexible, depending on the characteristics of the absorptions in the spectra and the identification goals. First, a set of segments is selected from a library spectrum, and then, the corresponding portions in a test spectrum are matched to compute correlation indices. The correlation indices of different segments are associated with weights related to a measure of the shape of the absorptions. A matching index is calculated by summing all the weighted correlation indices. The weights can be derived from the absorption characteristics such as band-depth, full width at half maximum (FWHM), band-area, or the spectral length of the absorptions \cite{kumari2023fully}. Alternatively, a combination of these specifications can be used for the weights. Each weight is proportioned by the total weight on the spectra to create a standardized matching index.\par
\begin{figure}[!b]
    \centering
    \subfloat{\includegraphics[width=\linewidth]{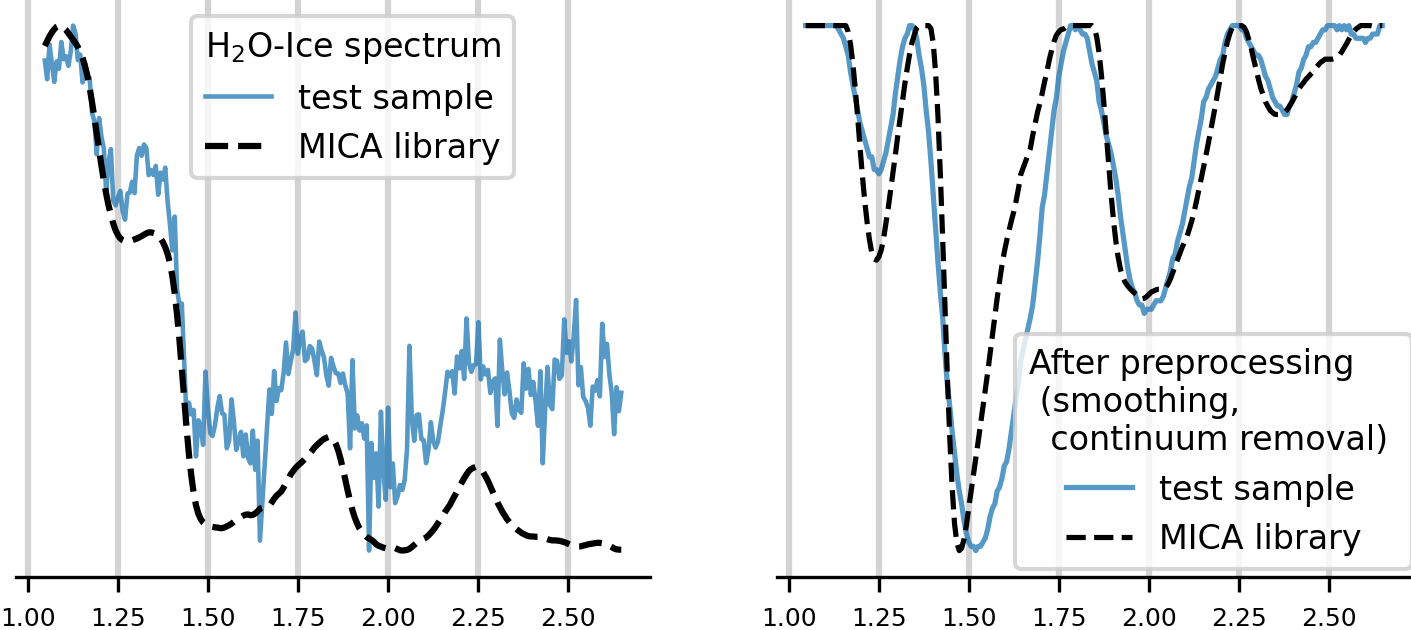}}\hfill
    \subfloat{\includegraphics[width=\linewidth]{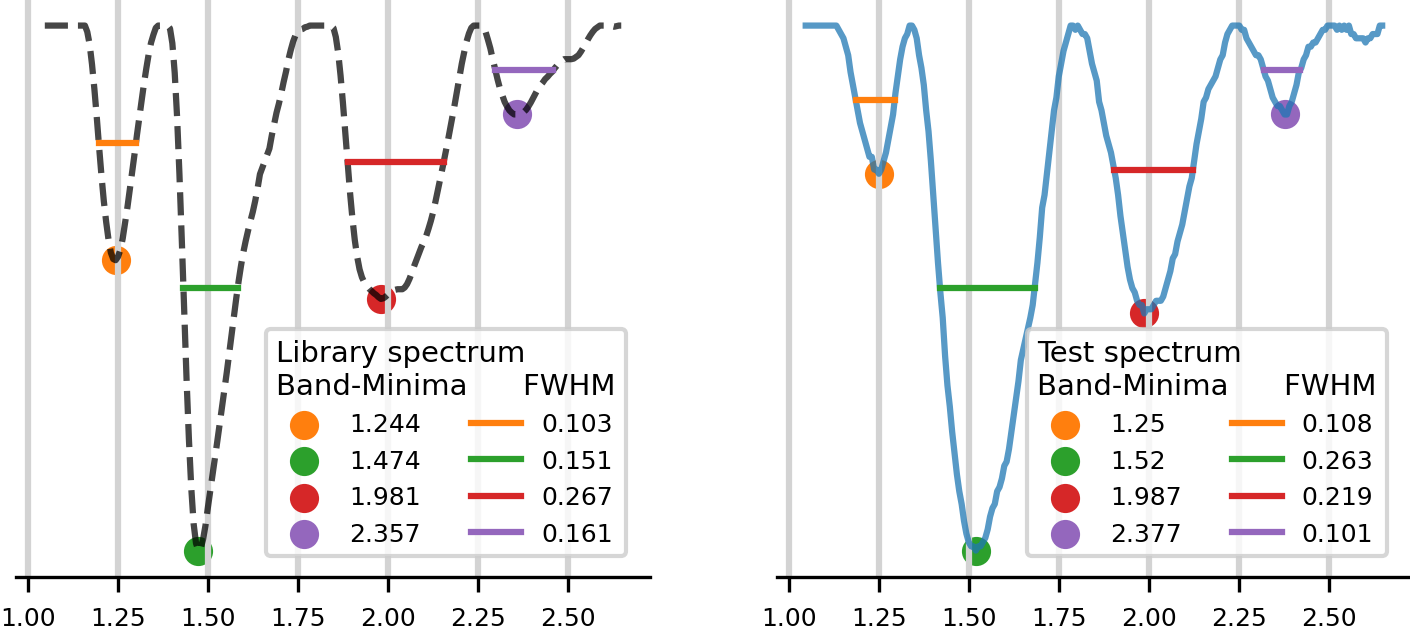}}\hfill
    \subfloat{\includegraphics[width=\linewidth]{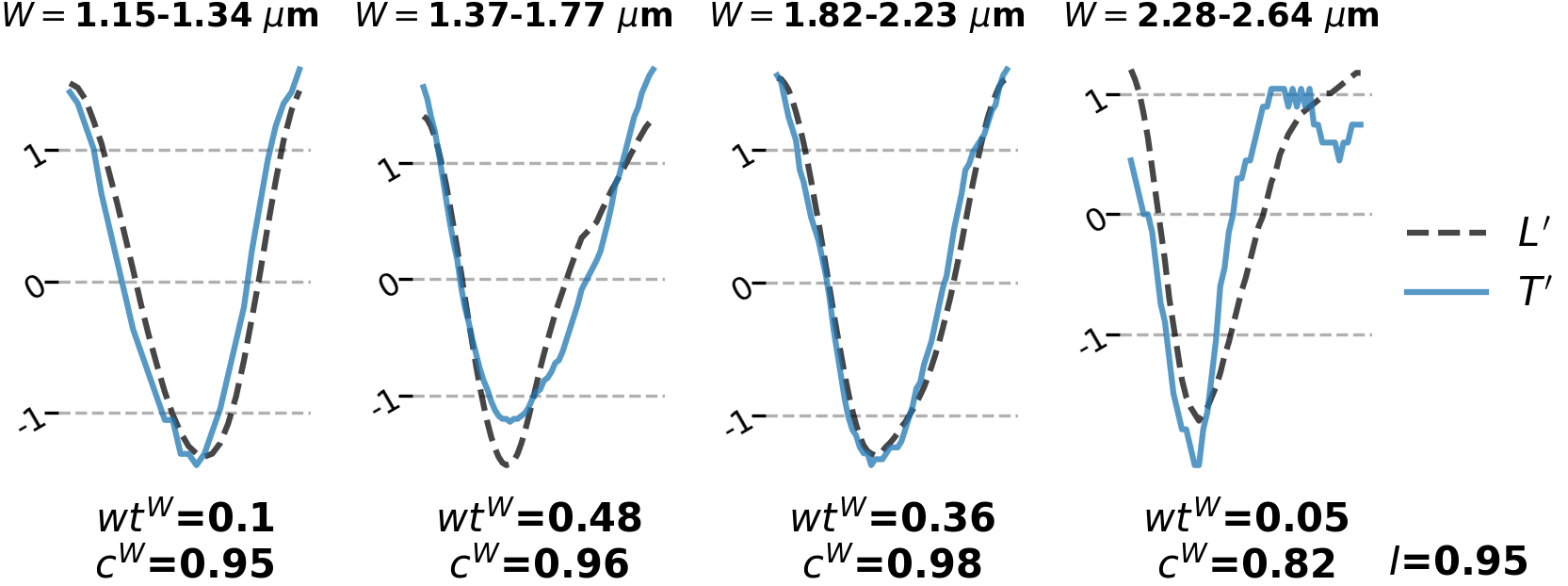}}\hfill
    \caption{
    {\bf Top Row: }A preprocessing pipeline 
    translates the two H\textsubscript{2}O-Ice spectra (from MICA spectral library \cite{viviano2014revised} and a CRISM MTRDR data) in the left image to the right image. 
    {\bf Middle Row: } Band-minima and FWHMs for all the prominent segments in the processed spectra.
    {\bf Bottom Row: } Correlation coefficients ($c^W$) of the spectra sgments and the matching index ($I$) between the two preprocessed spectra.
    }
    \label{fig:WSSC}
\end{figure}

For example, consider the approach to calculate the matching index followed in this study. Let, $\mathcal{W}$ be the set of distinct wavelength segments used to calculate the matching index. Let $f^{ W}$ and $d^{ W}$ respectively be the FWHM and the depth at the band-minima of such a segment $W\in\mathcal{W}$. The weight $wt^{ W}$ is defined as,
\begin{equation}
    wt^{ W}=\frac{f^{ W}*d^{ W}}{\sum\limits_{\forall W\in\mathcal{W}}(f^{ W}*d^{ W})}
\end{equation}
Considering $c^{ W}$ as the correlation index of the segment $W\in\mathcal{W}$, the matching index between the library and test spectra can be defined as,
\begin{equation}\label{eq:I}
    I=\sum_{\forall W\in\mathcal{W}}(wt^{ W}*c^{ W})
\end{equation}
A test spectrum is identified as the same class of library spectrum with which it has the highest matching index. Figure \ref{fig:WSSC} depicts a demonstration of calculating $I$. The $c^{ W}$'s in eq. \ref{eq:I} can be constrained to ensure non-negative values, i.e., not penalizing the non-correlated segments, which can contribute to correctly identifying mineral groups, if not the mineral class.\par
\vspace{.1cm}
{\bf How to choose the segments?}\par
The segments for examining specific absorption features can be determined either through prior knowledge about different absorption features or through an automated process, such as the apparent continuum removal method described in \cite{kumari:hal-04104263}. Figure \ref{fig:WSSC} demonstrates this process, where a H\textsubscript{2}O-Ice spectrum from the MICA library \cite{viviano2014revised} is preprocessed and four prominent absorptions are extracted. A test spectrum undergoes the same preprocessing to enhance its spectral signatures. For each absorption extracted from the library spectrum, the corresponding segment in the preprocessed spectrum is considered to calculate the correlation indices.
\section{Application on Mineral Identification}
\label{sec:ResultsAVRIS}
\subsection{AVIRIS data \cite{green1998imaging} (Earth surface)}
\label{subsec:ResultsAVRIS}
\begin{figure*}[!t]
    \centering
    \subfloat[{\scriptsize Ground truth}]{\includegraphics[width=.175\linewidth]{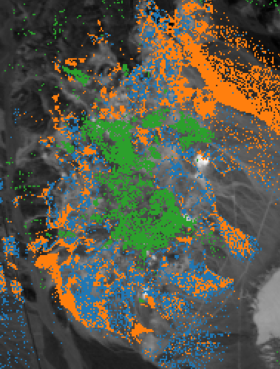}}\hfill
    \subfloat[{\scriptsize Cosine Similarity}]{\includegraphics[width=.175\linewidth]{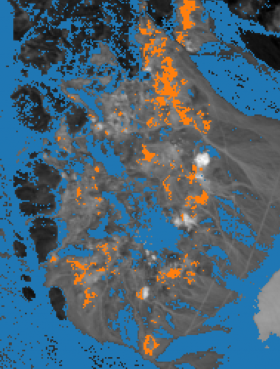}}\hfill
    \subfloat[{\scriptsize Corr. Coefficient}]{\includegraphics[width=.175\linewidth]{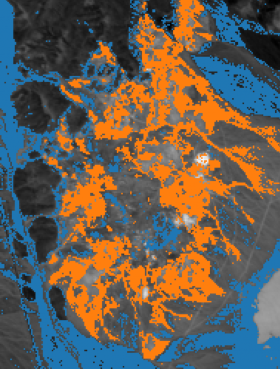}}\hfill
    \subfloat[{\scriptsize WSSC}]{\includegraphics[width=.175\linewidth]{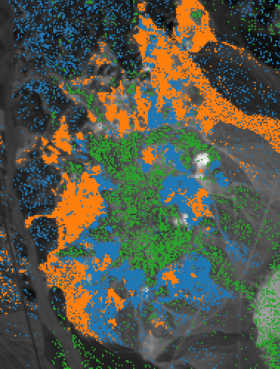}}\hfill
    \subfloat{
    \rotatebox{90}{
    \includegraphics[width=.15\linewidth]{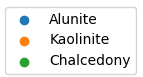}
    }
    }\hfill
    \caption{
    Some dominant minerals in the Alunite hill scene from AVIRIS Cuprite data. Ground-truth obtained from \cite{clark2003imaging}.
    }
    \label{fig:cuprite_result}
\end{figure*}

The Cuprite dataset, covering the area of Nevada, USA, can be acquired from the AVIRIS NASA website. The dataset comprised 224 channels, in the wavelength range of $370-2480$ nanometers combining the captures by four spectrometers. After eliminating the noisy channels and water absorption channels, 188 channels are left. A $250\times190$ pixel area containing the Alunite hill scene is used in this study. The preprocessing steps are applied to the spectral regions captured by the different spectrometers separately. Different sets of spectral regions are selected to identify different minerals. For instance, to identify alunite minerals, only the two ranges between $1.34$-$1.67 \,\mu m$ and $1.68$-$2.17 \,\mu m$, captured by spectrometers C and D, are considered, as these contain the key absorptions that distinguish alunite from similar minerals. Clark et al. proposed imaging spectroscopy \cite{clark2003imaging} for mineral classification, notably applied to Cuprite data, containing a vast set of ground truth. The spatial distribution maps of the most dominant minerals in the site, kaolinite, alunite and chalcedony, obtained by WSSC draw a very prominent visual similarity with the ground truth compared to other approaches, as shown in figure \ref{fig:cuprite_result}.\par
\subsection{CRISM data \cite{nasapds} (Martian Surface)}
\label{subsec:ResultsCRISM}
\begin{figure*}[!t]
    \centering
    \subfloat[{\scriptsize Browse Products}]{\includegraphics[width=.2\linewidth,height=.2\linewidth]{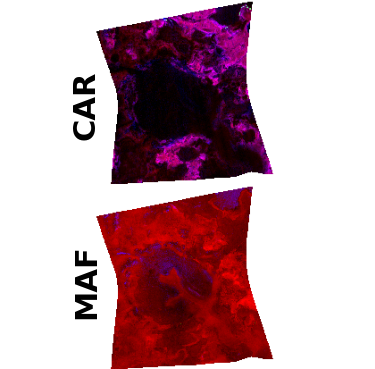}}\hfill
    \subfloat[{\scriptsize Cosine Similarity}]{\includegraphics[width=.175\linewidth,height=.175\linewidth]{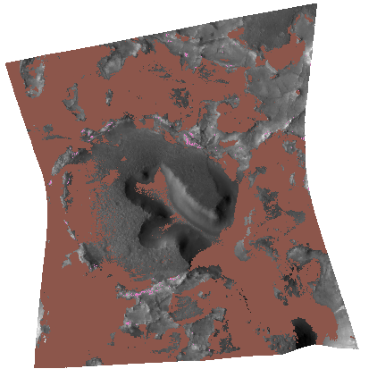}}\hfill
    \subfloat[{\scriptsize Corr. Coefficient}]{\includegraphics[width=.175\linewidth,height=.175\linewidth]{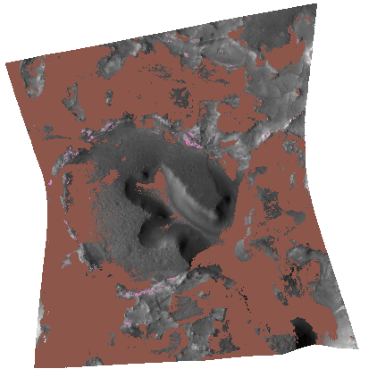}}\hfill
    \subfloat[{\scriptsize WSSC}]{\includegraphics[width=.175\linewidth,height=.175\linewidth]{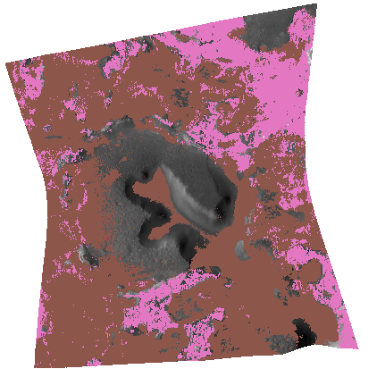}}\hfill
    \subfloat{
    \rotatebox{90}{
    \includegraphics[width=.175\linewidth]{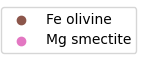}
    }
    }\hfill
    \caption{
    Dominant minerals in CRISM MTRDR data FRT3E12 from Nili Fossae region of Mars. In CAR browse product the distribution of Mg-Smectites is indicated in magenta, while in MAF the presence of Fe Olivines is indicated in red.
    }
    \label{fig:MTRDR_result}
\end{figure*}

MTRDR CRISM data is accessible via the NASA Planetary Data System. For this study, the MTRDR image FRT3E12 from the Nili Fossae region of Mars is considered which contains massive lowlands of the surface in $828\times843$ pixels each having 243 channels in the selected wavelength range of $1$-$2.6\,\mu m$. The matching segments are automatically extracted from the MICA spectral library \cite{viviano2014revised} for this case study. The dominant minerals in the scene, Fe Olivine and Mg Smectite deposits, are identified by WSSC which are aligned with the mineral distribution maps given in the browse products, as shown in figure \ref{fig:MTRDR_result}. Fe Olivines exhibit a distinctive bowl-shaped absorption across the broad wavelength range of $1.0$-$1.7\,\mu m$, enabling accurate detection also by cosine similarity and correlation coefficient on the full spectral domain. In contrast, Mg Smectites are identified by narrow absorption at $1.4 \,\mu m$ and a doublet around $2.35\,\mu m$, along with a common absorption feature of the phyllosilicate mineral group at $1.9\,\mu m$ \cite{kumari2023mineral}. These nuanced absorptions in small spectral segments are not discerned by cosine similarity and correlation coefficient but are accurately detected by WSSC.
\section{Conclusion}
\label{conclusion}
A spectra-matching technique called Weighted Sum of Segmented Correlation is introduced in this study, which specifically matches only the most informative or distinguishable segments within a hyperspectral spectrum. Experimental results demonstrated that this segmented matching surpassed full spectrum matching methods such as cosine similarity or correlation coefficient.

\bibliographystyle{IEEEbib}
\bibliography{refs}

\end{document}